\pdfoutput=1

\documentclass[11pt]{article}

\usepackage[preprint]{acl}

\usepackage{times}
\usepackage{latexsym}
\usepackage{booktabs}
\usepackage{amsmath}
\usepackage[T1]{fontenc}

\usepackage[utf8]{inputenc}

\usepackage{microtype}
\usepackage{tabularx}
\usepackage{inconsolata}

\usepackage{graphicx}

\newcommand\blfootnote[1]{%
  \begingroup
  \renewcommand\thefootnote{}\footnote{#1}%
  \addtocounter{footnote}{-1}%
  \endgroup
}
\title{\textsc{Open-Theatre}: An Open-Source Toolkit for LLM-based\\ Interactive Drama}

\author{
Tianyang Xu$^{\Diamond 1}$,
Hongqiu Wu$^{\Diamond 2,3,4}$,
Weiqi Wu$^{2,3,4}$,
Hai Zhao$^{\dagger 2,3,4}$ \\
$^1$ UM--SJTU Joint Institute, Shanghai Jiao Tong University \\
$^2$ AGI Institute, School of Computer Science, Shanghai Jiao Tong University \\
$^3$ Key Laboratory of Shanghai Education Commission for Intelligent Interaction \\
\hspace{0.8em} and Cognitive Engineering, Shanghai Jiao Tong University \\
$^4$ Shanghai Key Laboratory of Trusted Data Circulation and Governance in Web3 \\
\texttt{\{johnnie.walker,wuhongqiu\}@sjtu.edu.cn}
}

\usepackage{soul} 
\sethlcolor{yellow}
\begin{document}
\maketitle

\blfootnote{$^\Diamond$ The authors contributed equally. \quad $^\dagger$ Corresponding author. \textsc{Open-Theatre} is available at \url{https://github.com/johnnie193/Open-Theatre}, with a demo video for setup.}

\begin{abstract}

\emph{LLM-based Interactive Drama} introduces a novel dialogue scenario in which the player immerses into a character and engages in a dramatic story by interacting with LLM agents.
Despite the fact that this emerging area holds significant promise, it remains largely underexplored due to the lack of a well-designed playground to develop a complete drama.
This makes a significant barrier for researchers to replicate, extend, and study such systems.
Hence, we present \textsc{Open-Theatre}, the first open-source toolkit for experiencing and customizing LLM-based interactive drama.
It refines prior work with an efficient multi-agent architecture and a hierarchical retrieval-based memory system, designed to enhance narrative coherence and realistic long-term behavior in complex interactions.
In addition, we provide a highly configurable pipeline, making it easy for researchers to develop and optimize new approaches.

\end{abstract}

\section{Introduction} 
Interactive drama \citep{ryan1997interactive,10.1145/1240624.1240847} offers a novel participatory storytelling paradigm where users engage as an in-story character. In LLM-based interactive drama, the in-story characters (known as NPCs) are simulated by large language models (LLMs) \citep{DBLP:journals/corr/abs-2303-08774,DBLP:journals/corr/abs-2310-06825,DBLP:journals/corr/abs-2407-21783}, and the storytelling is collaboratively generated through dialogues between users and LLM agents.
\begin{figure*}[h]
\centering
\includegraphics[width=1\textwidth]{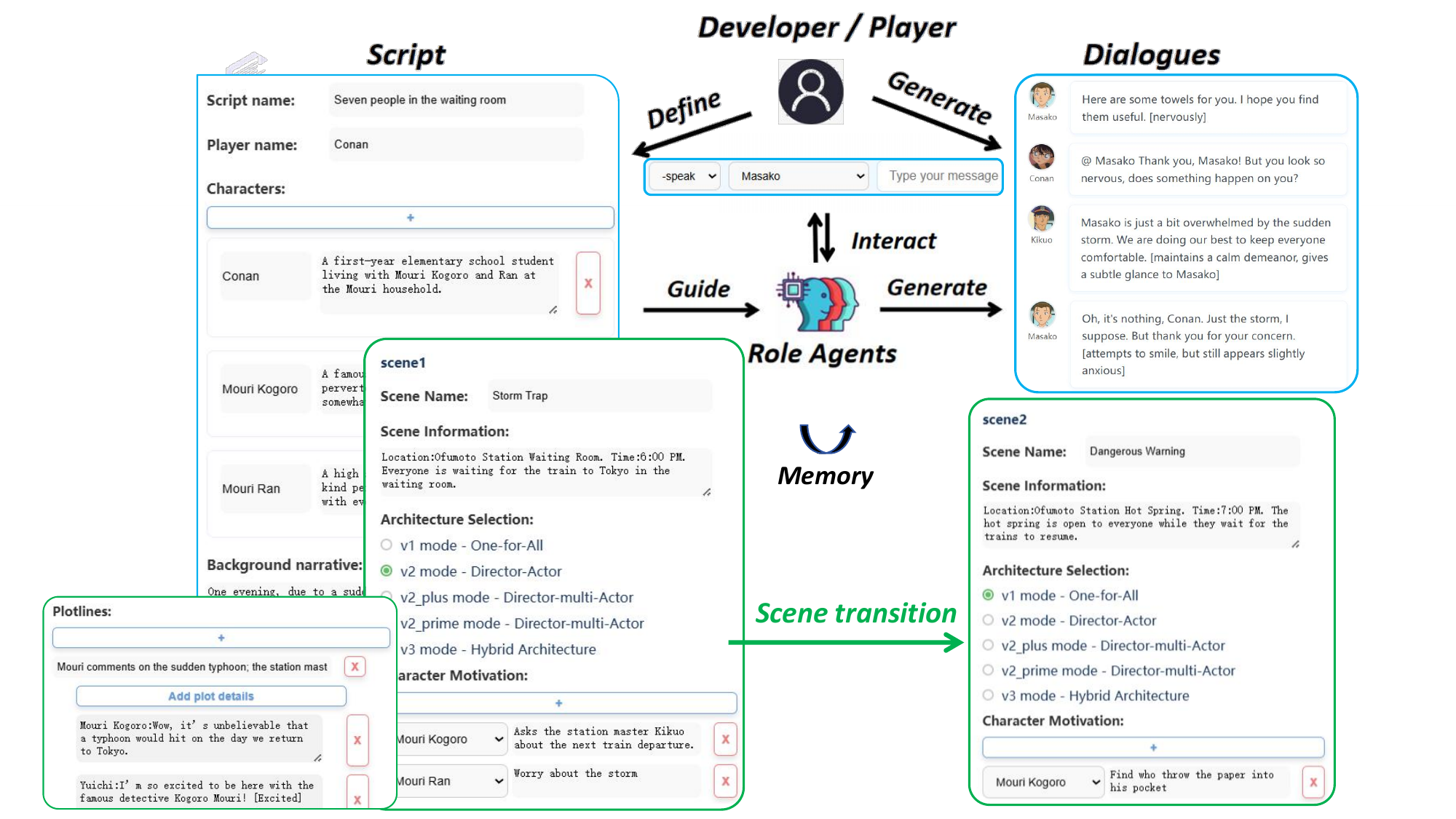}
\caption{A demonstration of LLM-based interactive drama using the interface of the Open-Theatre. The script is adapted from the popular anime \emph{Detective Conan}.}
\label{fig:main}
\end{figure*}

The incorporation of LLMs have significantly enhanced the realism and adaptability of the interactive experiences.
Despite growing interest in LLM-based interactive drama \citep{DBLP:books/daglib/0081917,DBLP:conf/acl/WuWJL0024,DBLP:conf/acl/HanCLXY24}, a lack of accessible, open-source tools persists, hindering researchers from easily creating, modifying, and experimenting with such interactive experiences.\footnote{A newest interactive drama: \url{https://github.com/gingasan/interactive-drama}}

To bridge this gap, we introduce \textbf{Open-Theatre}, an open-source toolkit designed for creating and experiencing configurable LLM-based interactive drama. 

Current interactive drama frameworks vary in their focus, with some prioritizing user freedom while others emphasize stronger narrative coherence. Open-Theatre integrates these different architectures, such as one-for-all, director-actor, and hybrid structures, allowing users to flexibly choose according to their narrative preferences. A key challenge, however, remains in balancing freedom and coherence, since excessive user influence may disrupt plot progression. To address this, Open-Theatre introduces the \textbf{Director-Global Actor} framework, an extension of the traditional Director-Actor setup, where a global actor agent enhances character decision-making and narrative coherence. Furthermore, Open-Theatre enhances long-term coherence through a novel \textbf{hierarchical memory management system}, enabling intelligent information recall and prioritization.


Through experimental evaluation across various scripts and architectures, this paper demonstrates how Open-Theatre effectively bridges users with the crafted narrative world. 

We outline the novel features of Open-Theatre:
\noindent$\bullet$ It is the first open-source toolkit that provides a highly configurable platform for creating and interacting with LLM-based interactive drama, enabling easy scripting, modification, and experimentation with dynamic narratives.

\noindent$\bullet$ It integrates and advances multiple architectures, utilizing sophisticated prompts and demonstrating the efficacy of the Director-Global Actor framework.


\noindent$\bullet$ It features an innovative hierarchical memory management system for role agents, promoting
human-like and consistent character responses.

\noindent$\bullet$ It allows secondary development for researchers with highly stable performance, enabling the integration of new architectures and customization of drama scripts and prompts in a lightweight manner.

\section{Related work}
\paragraph{Interactive Drama}
Computer-driven interactive drama has a long history of exploration \citep{DBLP:books/daglib/0081917}. More recently, \citet{DBLP:conf/acl/WuWJL0024} introduces the concept of LLM-based interactive drama, outlining its six key components and proposing a fine-tuning strategy for a global drama LLM. 
Building on this line of research, \citet{DBLP:conf/acl/WuWXZ025} propose a plot-based reflection mechanism that periodically reviews the evolving storyline and adjusts narrative elements to facilitate player-centered story curves.
    
\paragraph{Simulate Dramatic Characters by LLM Agents} 
LLM-based role-playing has seen significant advancements in generating dynamic and consistent character behavior \citep{chen2024socialbenchsocialityevaluationroleplaying,chen-etal-2023-large,NEURIPS2024_5875aca1,Zhou2023CharacterGLMCC,DBLP:conf/acl/WuW025}. Memory has been recognized as a core component of LLM-based agents, fundamentally shaping their ability to maintain coherence and adapt to long-term interactions. While prior work \citep{park2023generativeagentsinteractivesimulacra} has explored character memory over interactions, these systems often lack dynamic, context-aware memory management crucial for evolving dramatic narratives. General memory systems such as Mem0 \citep{chhikara2025mem0} and MemBank \citep{zhong2023memorybankenhancinglargelanguage} offer structured storage but struggle with customization based on plot progression. Open-Theatre addresses these limitations by incorporating a novel hierarchical memory system.

\paragraph{Multi-Agent Framework} 
Multi-agent systems \citep{tao2024magis,islam2024mapcodermultiagentcodegeneration} such as BabyAGI \citep{nakajima2023babyagi}, AutoGen \citep{wu2023autogenenablingnextgenllm} and Camel-AI \citep{li2023camelcommunicativeagentsmind}, aim to achieve complex and goal-oriented behavior. 
In the context of interactive narratives, multi-agent frameworks such as director-actor \citep{DBLP:conf/acl/HanCLXY24,DBLP:conf/acl/WuWXZ025} and ego-superego \citep{magee2024dramamachinesimulatingcharacter} coordinate agents to enable natural language interactions and strategic decision-making.
However, maintaining narrative coherence in open-ended interactions remains a key challenge, as character autonomy can lead to inconsistent or fragmented storylines. Our Open-Theatre consolidates and extends these existing architectures through the introduction of a Director–Global Actor framework. 

\section{LLM-based Interactive Drama}
We introduce the background of \textit{LLM-based Interactive Drama}.
It is a novel dialogue scenario where players experience an unfolding dramatic story by having dialogues with LLM-based characters.

As illustrated in Figure \ref{fig:main}, a \textbf{drama script} guides the entire dialogues, which is in an episodic structure of scenes. Each scene details its background, character motivations, and plotlines, dictating character behaviours and plot objectives. For instance, Scene 1 (\emph{Storm Trap}) depicts a scene where characters are trapped in the station waiting room due to a sudden typhoon.  Completing Scene 1's plotlines transitions to Scene 2 (\emph{Dangerous Warning}).

The drama script forms the foundation, guiding \textbf{role agents} (LLM agents simulating characters) to follow a specific architecture to collaborate together. Players could interact with any character in the ongoing scene, and characters would respond based on their motivations and plotlines defined in the script, adapting to player influence. For instance, in Figure \ref{fig:main}, a player (as \emph{Conan}) interacts with \emph{Masako}, who, despite being scripted to distribute towels, adaptively responds when questioned about her nervousness, balancing plotline adherence with player-driven interaction.


\section{Open-Theatre}
This section introduces our \textsc{Open-Theatre} toolkit designed for a highly configurable and easy-to-use LLM-based interactive drama. The core components of the Open-Theatre interface includes developer console, player console, and monitor. 

\subsection{Developer Console}
The Developer Console serves as the backbone of Open-Theatre, providing creators with tools for pre-configuring narratives and refining storytelling during interaction.

\paragraph{Script Management} This module enables comprehensive customization of the drama script, from foundational background settings to individual scene design. The \textbf{Background} module defines the narrative context, including setting, themes, and detailed character profiles. \textbf{Scenes} are broken down into discrete components that shape story progression: \textbf{Scene Information} defines location, time, and background; \textbf{Character Motivation} outlines scene-specific roles and motivations, ensuring agent alignment with narrative demands; \textbf{Plotlines} organize the scene story into manageable segments, ensuring logical progression and coherence and allowing predefinition of dialogues or events. The system supports real-time modifications to the script, enhancing dynamic narrative control.

\paragraph{Prompt Management}
While default prompts are pre-designed for general use,  users can refine them to suit specific narrative needs or subsequent development.

\paragraph{Save and Load Management}
This function enables persistent storage of the current script state and chat history, allowing revisiting specific narrative points and ensuring continuity across sessions. For researchers, it provides robust support for debugging, scenario testing, interaction replication, and iterative storytelling and replayability.

\subsection{Player Console}

The Player Console serves as the primary interface for players to engage in an interactive drama, enabling them to execute specific commands, trigger scene transitions, and influence the story's direction. Players, as characters, can select other characters for \textbf{interaction} or choose \textbf{autonomous drama progression}. Advanced features include \textbf{withdrawing} to experiment with responses and \textbf{navigating directly to the next or previous scene}. All character dialogues, user inputs, and system feedback are presented in real-time, offering greater flexibility and dynamic story exploration.

\subsection{Monitor}
The Monitor provides players and developers with essential real-time contextual details of the Open-Theatre system's processes, facilitating understanding and guiding adjustments to the script and prompts. It comprises several key components:

$\bullet$ \textbf{Current Script:} Offers a real-time view of the ongoing script's state and progression, allowing users to track drama development.

$\bullet$ \textbf{Characters:} Enables users to explore details about characters in the current scene, including their profile, memory, motivation and agent reaction if applicable.

$\bullet$ \textbf{System Feedback:} Displays the system's generated responses and real-time prompts, offering transparency into underlying processes and insights into system decisions for refinement.

$\bullet$ \textbf{Record:} Captures the complete state of the drama, including all major events, actions, and interactions throughout the narrative.

\subsection{Architectural Integration and Innovation}
The Open-Theatre toolkit enhances flexible narrative experiences by integrating existing architectures and introducing its own novel designs. 

\paragraph{One-for-All}
This architecture uses a single global agent to control all characters and the progression of drama. It is computationally efficient and suitable for scenes that prioritize narrative flow over player agency, such as expository or background-setting scenes.
\paragraph{Director-Actor}
In this architecture, each character is represented by an independent actor agent, while a higher-level director agent coordinates and guides the narrative. It allows for complex decision-making, making it ideal for scenes requiring interactive depth. However, it often incurs significant computational cost and latency due to requiring individual LLM requests for each actor's decision-making and profile adherence.

\paragraph{Hybrid Architecture}
This architecture switches between the director-actor and one-for-all architectures based on the scene’s characteristics, ensuring a balance between narrative immersion and interactive depth.

\paragraph{Director-Global Actor} We extend the traditional Director-Actor architecture, employing a single, centralized global actor agent that makes decisions for all characters. This global actor agent serves as a high-level consciousness for the entire dramatic ensemble. Guided by the director agent, it synthesizes comprehensive, real-time knowledge of all characters' profiles, memories, and states, providing a unifying context individual actor agents would otherwise lack. This capability not only ensures more strategically aligned and cohesive character actions, preventing contradictions and maintaining consistent character arcs, but also drastically improves efficiency, reducing redundant LLM API calls and latency for individual actor agents.
\section{Hierarchical Memory System}

Open-Theatre introduces a novel hierarchical memory architecture to provide narrative agents with a persistent, contextually-aware memory for extended, interactive narratives. While previous works \citep{park2023generativeagentsinteractivesimulacra} emphasized recency, importance, and relevance in memory recall, our model extends this by introducing a partitioned memory space and a dynamic scoring model tailored for dramatic agency. This adaptive memory lifecycle ensures computational tractability and psychological plausibility throughout extended interactions.

\begin{figure*}[h]
    \centering
    \includegraphics[width=0.8\textwidth]{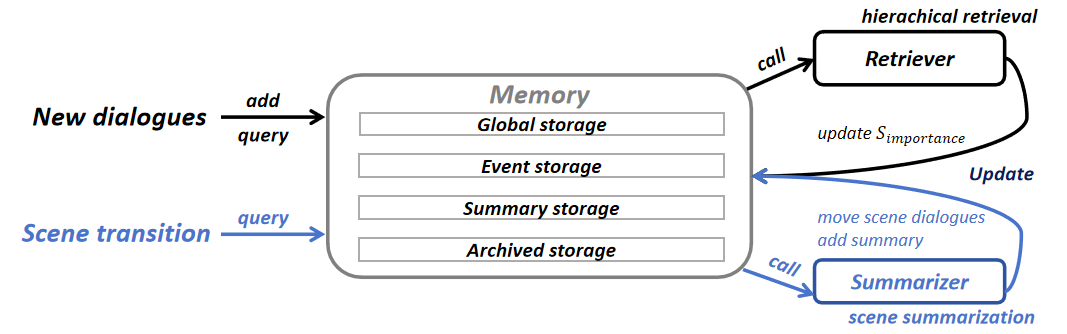} 
    \caption{\textbf{Hierarchical Memory Architecture.} New dialogues are ingested into the Event Store and call retriever to query all stores.The scene transition signal calls the summarizer module to create new entries in the Summary Store and move original records to the Archive Store, completing the memory lifecycle.}
    \label{fig:architecture}
\end{figure*}
\subsection{Architectural Design}
The architecture comprises four specialized memory stores, structured to model the varying cognitive accessibility of different information types within a partitioned memory space. The fundamental unit is a memory entry, a natural language description with metadata including semantic type, scene identifier, and a dynamically updated importance score.

These stores include: a \textbf{Global Store} for foundational, static knowledge like core identity attributes; an \textbf{Event Store} capturing the chronological stream of episodic experiences including dialogues and actions; a \textbf{Summary Store} housing condensed, abstractive summaries of events produced by memory consolidation; and an \textbf{Archive Store} serving as a long-term repository for consolidated, less frequently accessed records, preserving historical context for potential future retrieval.
\subsection{Retriever}
To manage the cognitive load of long dialogues, a retriever module computes relevant information from the memory stores in response to new dialogue turns. For a given new dialogue query $q$, the final retrieval score for each candidate memory entry $c$ is computed based on three factors: relevance, importance, and narrative recency, defined as:
\begin{equation}
\label{eq:main_score}
S_{final}(c) = P_{recency} \cdot (S_{relevance} + S_{importance})
\end{equation}
\subsubsection{Base score}
The base retrieval score is an additive combination of relevance and importance. The \textbf{Relevance Score ($S_{relevance}$)} is computed using a standard hybrid method, fusing lexical signals from BM25Okapi with semantic similarity from FAISS-indexed sentence embeddings. For sentence embeddings, we utilize the all-MiniLM-L6-v2 model. The system processes the narrative at the character utterance level; each dialogue turn is treated as a single memory entry. These entries are aggregated into larger chunks for efficient storage and retrieval. Chunking is performed with a configurable size (\texttt{chunk\_max\_pieces} = 5) and an overlap (\texttt{chunk\_overlap\_pieces} = 1), allowing the system to preserve local coherence while reducing fragmentation. The \textbf{Importance Score ($S_{importance}$)} is an evolving score unique to each memory entry. It increases with each retrieval event, allowing the system to learn a memory's prominence from its usage history. The relative weighting between relevance and importance is determined by the \texttt{ADDITION\_WEIGHT} (default set to 0.05), ensuring that the retrieval is primarily driven by immediate semantic and lexical relevance, while the learned importance serves as a subtle long-term signal to capture memory salience.  This design ensures frequently recalled memories can influence agent behavior more, consistent with cognitive theories of memory salience.

\subsubsection{Recency Punishment}
The base score is modulated by a dynamic factor ($P_{\text{recency}} \in (0, 1]$) that penalizes temporally distant memories. Our model implements a two-tiered decay logic to differentiate narrative recency types:

\paragraph{Inter-Scene Penalty} 
To model the decay of
relevance \textit{across} contextual shifts, for memory entries from a different scene ($c_{\text{scene}} \neq S_{\text{current}}$), a penalty based on scene distance is applied:
    \begin{equation}
    P_{\text{recency}} = (1 + \alpha \cdot |S_{\text{current}} - c_{\text{scene}}|)^{-1}
    \end{equation}
    where $\alpha$ is a decay constant, empirically tuned to control the rate of forgetting across scene transitions, set to 0.25 as default.

\paragraph{Intra-Scene Dialogue Penalty} 
For records \textit{within} the current scene, a penalty based on turn order captures the immediate conversational context:
    \begin{equation}
    P_{\text{recency}} = (1 + \beta \cdot \text{TurnsAgo})^{-1}
    \end{equation}
    where $\beta$ is a per-turn decay constant, and \textit{TurnsAgo} represents number of dialogue turns since creation. We set $\beta$ to 0.005, ensuring that recent turns are slightly favored but older relevant memories within the scene are not entirely excluded. For other memory types within the current scene, $P_{\text{recency}}$ is $1.0$, incurring no penalty.

\subsection{Summarizer}
To manage the cognitive load inherent in long narratives, the system implements a summarizer module for memory consolidation. It processes records from the Event Store for completed scenes, leveraging an LLM to generate abstractive summaries that are added to the Summary Store. This process enriches the agent's memory base with high-level, semantic knowledge from raw episodic experiences, concurrently reducing the pressure to retain all original records, which are then moved to the Archive Store for long-term preservation.

\section{Evaluation}

We conduct a rigorous evaluation to assess the performance of distinct agent architectures with and without memory system. Our analysis is performed across three diverse scripts to ensure robust, genre-agnostic findings: a detective story (adapted from \textit{Detective Conan}), an adventure tale (adapted from \textit{Harry Potter}), and a classical drama (adapted from \textit{Romeo and Juliet}).

\begin{table*}[t]
\centering
\setlength\tabcolsep{5pt} 
\caption{A multi-dimensional assessment of different architectures with and without memory configurations. Qualitative scores are averaged across three diverse scripts and rated on a 1-5 scale by our AI Judge.}
\label{tab:arch_comparison}
\begin{tabular}{lccccccccc} 
\toprule
\textsc{Architecture} & \textsc{Config} & \multicolumn{3}{c}{\textsc{Memory}} & \multicolumn{3}{c}{\textsc{Architecture}} & \multicolumn{2}{c}{\textsc{Efficiency}} \\
\cmidrule(lr){3-5}\cmidrule(lr){6-8}\cmidrule(lr){9-10}
& & \textbf{RA} & \textbf{RP} & \textbf{NC} & \textbf{CC} & \textbf{MAC} & \textbf{PA} & \textbf{Latency (s)} & \textbf{LLM Calls} \\
\midrule
\textit{One-for-All} & w/o RAG & / & 3.8 & 3.8 & 3.5 & 4.4 & 4.3 & \textbf{12.1} & \textbf{1.0} \\
& w/ RAG & 4.6 & 4.4 & 4.2 & 3.4 & 4.4 & \textbf{4.6} & 16.8 & 1.0 \\
\midrule
\textit{Director-Actor} & w/o mem & / & 4.0 & 3.7 & \textbf{4.6} & 3.3 & 3.7 & 34.5 & 3.4 \\
& w/ mem & 4.5 & \textbf{4.7} & 4.0 & \textbf{4.6} & 3.4 & 4.0 & 40.6 & 3.4 \\
\midrule
\textbf{\textit{Director-Global Actor}} & w/o mem & / & 3.9 & 4.0 & 4.1 & \textbf{4.5} & 4.0 & 20.5 & 2.0 \\
& w/ mem & 4.5 & 4.4 & \textbf{4.6} & 4.3 & \textbf{4.5} & 4.3 & 25.0 & 2.0 \\
\bottomrule
\end{tabular}
\end{table*}

\subsection{Experimental Setup}

To ensure a comprehensive and objective assessment, we employed an automated evaluation pipeline. We utilized a suite of AI agents to act as players, designed with 10 distinct personas (ranging from cooperative to aggressive) to test system robustness and
adaptability of each system configuration against a wide spectrum of behaviors. Following each simulated playthrough, we use an impartial Judge Agent (powered by GPT-4o) to score the complete interaction log, which provides scalable, consistent, and reproducible ratings based on defined metrics, effectively minimizing human bias and effort.

\subsection{Evaluation Metrics}

\subsubsection{Memory System Performance}
We evaluate our hierarchical memory system through three complementary dimensions:

\noindent$\bullet$ \textbf{Retrieval Accuracy (RA):} Success rate of retrieving relevant historical information.

\noindent$\bullet$ \textbf{Response Plausibility (RP):} Degree to which memory-based responses align well with the character and the current dialogue context.

\noindent$\bullet$ \textbf{Narrative Coherence (NC):} Long-term story consistency through memory integration.

\subsubsection{Architectural Performance}
We assess the interactive experience under different architectures through:

\noindent$\bullet$ \textbf{Character Consistency (CC):} Maintenance of distinct character personalities.

\noindent$\bullet$ \textbf{Multi-Agent Coordination (MAC):} Quality of inter-character strategic alignment.

\noindent$\bullet$ \textbf{Plot Adherence (PA):} Effectiveness in narrative progression and engagement.

\subsubsection{System Efficiency}
We measure computational performance through:

\noindent$\bullet$ \textbf{Response Latency:} Average response time per interaction turn (seconds/turn).

\noindent$\bullet$ \textbf{Computational Cost:} Number of LLM API calls required per turn.

\subsection{Results}

Table~\ref{tab:arch_comparison} presents our evaluation results, comparing different architectures with and without our hierarchical memory system.

\subsubsection{Memory System Validation}
Across all architectures, integrating the hierarchical memory system yields consistent improvements in response quality and long-term coherence. Specifically, \textbf{Retrieval Accuracy (RA)} exceeds \textbf{4.5} out of 5, confirming the system’s ability to retrieve relevant past information reliably. 
When compared to the baseline without memory system, \textbf{Response Plausibility} saw a substantial improvement (e.g., from \textbf{3.8} to \textbf{4.4} in One-for-all), indicating that memory-based responses are much better aligned with character and current dialogue context. These gains translate into higher \textbf{Narrative Coherence} (e.g., up to \textbf{4.6} in Director-Global Actor), showing improved consistency over long interactions. Even in simpler settings like One-for-All, memory boosts NC from \textbf{3.8} to \textbf{4.2}, further supporting the general utility of our memory system.

\subsubsection{Architecture Performance}
In terms of architectural design, the proposed \textbf{Director-Global Actor} architecture offers a strong balance between multi-agent coordination and narrative coherence, while maintaining competitive efficiency. Unlike Director-Actor, which excels at \textbf{Character Consistency (CC=4.6)} but suffers from lower \textbf{Multi-Agent Coordination (MAC=3.3)} and \textbf{Plot Adherence (PA=3.7)}, the Director-Global Actor achieves strong performance across almost all metrics when paired with our memory system. It scores highest on \textbf{NC (4.6)} and maintains high values for \textbf{MAC (4.5)} and \textbf{PA (4.3)}, suggesting that centralized reasoning leads to better plot alignment and action coordination among characters.

\subsubsection{System Efficiency}
In terms of efficiency, the One-for-All architecture requires only 1.0 LLM call and achieves the lowest latency. However, this simplicity comes at the cost of limited multi-character reasoning, resulting in lower coordination and consistency scores.
By contrast, while the Director-Actor architecture incurs higher overhead, requiring 3.4 LLM calls per turn due to decentralized decision-making, the Director-Global Actor strikes a more favorable balance. It reduces the number of calls to 2.0 while maintaining high-quality outputs, confirming its computational viability for complex narrative generation. The competitive latency offers a favorable trade-off between performance and runtime cost.



\section{Conclusion}
This paper proposed Open-Theatre, a toolkit designed to facilitate the creation and interaction for LLM-based interactive drama. It assembles and advances different architectures, notably proposing a Director-Global Actor architecture and integrating a novel hierarchical memory system. Our experimental evaluation demonstrates that Open-Theatre provides a flexible and configurable platform for users to experience in a lightweight manner and facilitates secondary development. We hope this toolkit contributes to the broader field of interactive storytelling and inspires future research in AI-assisted narrative development.

\section{Limitation}
While an AI judge offers a highly scalable and standardized method for evaluating user experience, it might not fully capture the nuanced and subjective elements of human perception. Certain subtleties that only direct human feedback can reveal could be overlooked.

Also, our validation relies mostly on GPT-4o (accessed on August 6, 2024), limiting insights into framework generalizability across diverse LLMs. Future work will prioritize hybrid evaluation strategies and cross-model benchmarking to address these gaps.

\bibliography{acl_latex}

\appendix



\end{document}